\documentclass[11pt,letterpaper]{article}
\usepackage{naaclhlt2016}
\usepackage{times}
\usepackage{latexsym}
\usepackage{amssymb}
\usepackage{geometry}
\usepackage{listings}
\usepackage{color}
\usepackage[usenames,dvipsnames,svgnames,table]{xcolor}
\usepackage[utf8]{inputenc}
\usepackage[T1]{fontenc}

\naaclfinalcopy 
\def\naaclpaperid{***} 

\title{INSIGHT-1 at SemEval-2016 Task 5: Deep Learning for Multilingual Aspect-based Sentiment Analysis}

\author{Sebastian Ruder\textsuperscript{1}\textsuperscript{2}
	\And
		Parsa Ghaffari\textsuperscript{2}
		\And
		John G. Breslin\textsuperscript{1}
		\AND
	    \normalfont{\textsuperscript{1}Insight Centre for Data Analytics}\\
	    National University of Ireland, Galway\\
	    {\tt firstname.lastname@insight-centre.org}
	    \AND
	    \normalfont{\textsuperscript{2}Aylien Ltd.}\\
	    Dublin, Ireland\\
	    {\tt firstname@aylien.com}
  }
\date{}

\begin{document}

\maketitle

\begin{abstract}
This paper describes our deep learning-based approach to multilingual aspect-based sentiment analysis as part of SemEval 2016 Task 5. We use a convolutional neural network (CNN) for both aspect extraction and aspect-based sentiment analysis. We cast aspect extraction as a multi-label classification problem, outputting probabilities over aspects parameterized by a threshold. To determine the sentiment towards an aspect, we concatenate an aspect vector with every word embedding and apply a convolution over it.
Our constrained system (unconstrained for English) achieves competitive results across all languages and domains, placing first or second in 5 and 7 out of 11 language-domain pairs for aspect category detection (slot 1) and sentiment polarity (slot 3) respectively, thereby demonstrating the viability of a deep learning-based approach for multilingual aspect-based sentiment analysis.
\end{abstract}

\section{Introduction}

With access to the Internet becoming more prevalent, an inreasing number of people express their opinions online in a plethora of languages. Sentiment analysis \cite{Liu2012} enables us to derive shallow insights from these opinions related to their overall polarity. Often, however, e.g. in reviews, people do not express their opinion towards the entity as a whole, but refer to specific aspects such as the service in a restaurant.

Aspect-based sentiment analysis allows us to go deeper and determine sentiment towards such aspects of an entity. Past research in aspect-based sentiment analysis has largely focused on the English language, while SemEval 2016 Task 5 \cite{SemEval2016:task5} for the first time provides a forum for multilingual aspect-based sentiment analysis.

Recently, deep learning-based approaches have demonstrated remarkable results for text classification and sentiment analysis \cite{Kim2014}. A cascade of non-linearities allows them to model complex functions such as sentiment compositionality, while their ability to process raw signals renders them language and domain independent. In spite of these factors, they have largely gone untested for aspect-based sentiment analysis, particularly in the multilingual setting.

In this paper, we introduce our deep-learning based approach to aspect-based sentiment analysis as part of our participation in SemEval-2016 Task 5 Aspect-based Sentiment Analysis Slot 1 (Aspect Category Detection) and Slot 3 (Sentiment Polarity) .

\section{Related work}

Aspect-based sentiment analysis is traditionally split into an aspect extraction and a sentiment analysis subtask.

Previous approaches to aspect extraction framed the task as a multiclass classification problem and relied mostly on CRS that leveraged a plethora of common features, e.g. NER, POS tagging, parsing, semantic analysis, bag-of-words, as well as domain-dependent ones, such as word clusters learnt from Amazon and Yelp data, while previous sentiment analysis approaches have used different classifiers with a wide range of features based on n-grams, POS, negation words, and a large array of sentiment lexica \cite{Pontiki2014a,Pontiki2015}.

Past deep learning-based approaches have focused mostly on the sentiment analysis subtask: Tang et al. \shortcite{Tang2015} use a target-dependent LSTM to determine sentiment towards a target word, while Nguyen and Shirai \shortcite{Nguyen2015a} use a recursive neural network that leverages both constituency as well as dependency trees.

In contrast to previous approaches, our model neither relies on expensive feature engineering, availability of a parser, nor positional information, but solely on a language's input signals.

\section{Model} \label{sec:model}

The model architecture we use is an extension of the CNN structure used by Collobert et al. \shortcite{Collobert2011a}, which has been successfully used by many others \cite{Kim2014}.

The model takes as input a text, which is padded to length $n$. We represent the text as a concatentation of its word embeddings $x_{1:n}$ where $x_i \in \mathbb{R}^k$ is the $k$-dimensional vector of the $i$-th word in the text.

The convolutional layer slides filters of different window sizes over the input embeddings. Each filter with weights $w \in \mathbb{R}^{hk}$ generates a new feature $c_i$ for a window of $h$ words according to the following operation:

\begin{equation} \label{eq:featuremap}
c_i = f(w \cdot x_{i:i+h-1} + b) 
\end{equation}

Note that $b \in \mathbb{R}$ is a bias term and $f$ is a non-linear function, ReLU \cite{Nair2010} in our case. The application of the filter over each possible window of $h$ words or characters in the sentence produces the following feature map:

\begin{equation} 
c = [c_1, c_2, ..., c_{n-h+1}]
\end{equation}

Max-over-time pooling in turn condenses this feature vector to its most important feature by taking its maximum value and naturally deals with variable input lengths.

A final softmax layer takes the concatenation of the maximum values of the feature maps produced by all filters and outputs a probability distribution over all output classes.

\section{Methodology}

   \lstset{
    language=xml,
    tabsize=1,
    caption=Example sentence with aspect and sentiment annotations for the English laptops domain.,
    label=lst:example_sentence,
    frame=shadowbox,
    rulesepcolor=\color{gray},
    xleftmargin=20pt,
    framexleftmargin=15pt,
    keywordstyle=\color{blue}\bf,
    commentstyle=\color{OliveGreen},
    stringstyle=\color{red},
    numbers=left,
    numberstyle=\tiny,
    numbersep=5pt,
    breaklines=true,
    showstringspaces=false,
    basicstyle=\small,
    emph={food,name,price},emphstyle={\color{magenta}}}    

\begin{lstlisting}[float=*]
<sentence id="347:0">
    <text>I bought it for really cheap also and its AMAZING.</text>
    <Opinions>
        <Opinion category="LAPTOP#PRICE" polarity="positive"/>
        <Opinion category="LAPTOP#GENERAL" polarity="positive"/>
    </Opinions>
</sentence>
\end{lstlisting}

\subsection{Preprocessing}

We lower-case and tokenize the corpus, where applicable, keeping the 10,000 most frequent words as the vocabulary for each language and domain. For Chinese, in preparation for the previous step, we additionally segment the corpus using the \texttt{mmseg} Python library. 

\subsection{Hyperparameters}

We randomly split off 20\% of each training data set as a validation set. We use this to optimize hyperparameters via random search over a wide range of values.

For both tasks and all languages and domains, we use the following hyperparameters, which are similar to those reported by Kim \shortcite{Kim2014}: mini-batch size of 10, maximum sentence length of 100 tokens, word embedding size of 300, dropout rate of 0.5, and 100 filter maps. We use filter lengths of 3, 4, and 5, and of 4, 5, and 6 for aspect extraction and aspect-based sentiment analysis respectively since these produced good results for the respective task.

English word embeddings are initialized with 300-dimensional GloVe vectors \cite{Pennington2014} trained on 840B tokens of the Common Crawl corpus for the unconstrained submission. Word embeddings for the constrained submission, for all other languages, as well as for words not present in the pre-trained set of words are initialized randomly. 

We train for 15 epochs using mini-batch stochastic gradient descent, the Adadelta update rule \cite{Zeiler2012}, and early stopping.

\subsection{Aspect Category Detection}

To extract aspects, e.g. \texttt{LAPTOP\#PRICE} and \texttt{LAPTOP\#GENERAL} from sentences as in Listing \ref{lst:example_sentence}, we cast aspect extraction as a multi-label classification problem and train a convolutional neural network (CNN) to output probability distributions over aspects, minimizing the cross-entropy loss. To model multi-label output as a probability distribution, we define an aspect $a$'s probability $p$ given a sentence $s$ as $p(a | s) = 1 / n$ if $a$ appears in $s$ and $s$ contains $n$ aspects, otherwise $p(a | s) = 0$. We define a threshold $f$ and discard all aspects with $p(a | s) < f$. After training, we select $f$ maximizing the F1 score on the validation set.

We observe that aspect distributions vary significantly depending on the domain and language. For instance, the English laptops domain contains 82 aspects, while the restaurants domain only contains 13 aspects.

In every domain, we thus replace all aspects that occur less than 5 times with an \texttt{OTHER} aspect.\footnote{We found that replacing all aspects with fewer than 5 occurrences yields the best trade-off between accuracy and recall.} E.g. this produces 51 aspects covering 98\% of occurrences and all 13 aspects for the English laptops and restaurants domain respectively. 
For instance, in the English laptops domain, aspects such as \texttt{HARDWARE\#MISCELLANEOUS} and  \texttt{BATTERY\#USABILITY}, which occur less than 5 times are replaced with \texttt{OTHER} during training. We add a \texttt{NONE} aspect to each sentence containing no aspect to enable the CNN to make no aspect prediction. During inference, every time the model predicts \texttt{OTHER}, the most frequent aspect replaced by \texttt{OTHER} for each domain is output instead. For the English laptops domain, this can be one of several aspects, e.g. \texttt{SOFTWARE\#QUALITY} occurring four times. Finally, we discard all predicted \texttt{NONE} aspects.

We experimented with producing representations for the preceding and subsequent sentence to take context information into account, but this did not improve results.

\subsection{Sentiment Polarity}

For aspect-based sentiment analysis, we feed the aspect vector together with the word embeddings of the input sentence into a CNN. To obtain the aspect vector, we follow an approach similar to the one used by Socher et al. \shortcite{Socher2013} to represent named entities: We split each aspect into its constituent tokens, e.g. \texttt{RESTAURANT\#GENERAL} $\rightarrow$ \emph{restaurant, general}. We embed the tokens of all aspects in an embedding space. We then look up the embedding of each of the tokens and average them to retrieve the aspect vector. This way, the model should learn that aspects sharing the same entity, e.g. \emph{restaurant} are correlated without the need to train several tiered models to classify between entities (\emph{restaurant}) and attributes (\emph{general}). 

For aspect-based sentiment analysis in the English language, we embed aspect tokens in the same embedding space as word tokens to exploit the semantics of pre-trained embeddings. For all other languages, we keep the embedding spaces separate, as aspect tokens are in English and word tokens are in the respective languages. Translating aspect tokens into the source language did not provide any benefits, but the use of pre-trained embeddings in the source language could ameliorate this.

We have experimented with different variants of adding aspect embeddings to our model: We evaluated summation, concatenation, and multiplication of word vectors and aspect vectors before the convolution, and multiplication and concatenation of the max-pooled sentence vector and the aspect vector after the convolution. We find that concatenating each word vector with the aspect vector before the convolution yields the best results.

To summarize, for the sentence in Listing \ref{lst:example_sentence} and the aspect \texttt{LAPTOP\#PRICE}, our model first pads the input sentence, then looks up the embeddings of each of the input words. It creates the aspect vector by splitting \texttt{LAPTOP\#PRICE} into the aspect tokens \emph{laptop} and \emph{price}. For these, it looks up the embeddings in the aspect embedding space (which is the same as for word embeddings for English) and averages both embeddings. The resulting aspect vector is then concatenated with each word vector, which are then concatenated to produce a 100x600 sentence matrix. Convolutions, max-pooling and softmax are applied to this matrix as described in section \ref{sec:model}.

We observe for some languages an incremental performance improvement when using additional average-pooling as reported by Tang et al. \shortcite{Tang2014a}. We further note that simply using a low-dimensional embedding space to embed aspects leads to superior results on a few occasions when no pre-trained word embeddings are used.

\section{Evaluation}

We have participated for all domains and languages in Slot 1: Aspect Category Detection and Slot 3: Sentiment Polarity. We report results for aspect extraction in Table \ref{tab:aspect_extraction} and results for aspect-based sentiment analysis in Table \ref{tab:sentiment_analysis}.

\subsection{Aspect Category Detection} \label{sec:aspect_extraction}

\begin{table}[]
\centering
\begin{tabular}{r l | c | c | c}
\textbf{Lg.} & \textbf{Dom.} & \textbf{F1} & \textbf{Top F1} & \textbf{R.} \\\hline
EN & REST & \begin{tabular}[c]{@{}l@{}}68.108 \textbf{U}\\ 58.303 \textbf{C}\end{tabular} & 73.031 & 9/30 \\
SP & REST & 61.370 & 70.588 & 5/9 \\
FR & REST & 53.592 & 61.207 & 4/6 \\
RU & REST & 62.802 & 64.825 & 2/7 \\
DU & REST & 56.000 & 60.153 & 2/6 \\
TU & REST & 49.123 & 61.029 & 5/5 \\
AR & HOTE & 52.114 & 52.114 & 1/3 \\
EN & LAPT & \begin{tabular}[c]{@{}l@{}}45.863 \textbf{U}\\ 41.458 \textbf{C}\end{tabular} & 51.937 & 10/22 \\
DU & PHNS & 45.551 & 45.551 & 1/4 \\
CH & CAME & 25.581 & 36.345 & 2/4 \\
CH & PHNS & 16.286 & 22.548 & 3/4
\end{tabular}
\caption{F1 and rank of our system for aspect extraction for each language and domain in comparison to the best system. Lg.: Language. Dom.: Domain. R.: Rank. EN: English. SP: Spanish. FR: French. RU: Russian. TU: Turkish. AR: Arabic. DU: Dutch. CH: Chinese. REST: Restaurants. HOTE: Hotels. LAPT: Laptops. PHNS: Phones. CAME: Cameras. U: Unconstrained submission. C: Constrained submission.}
\label{tab:aspect_extraction}
\end{table}

Despite using only the input sentence as data, our system is able to achieve competitive performance for multilingual aspect extraction, placing first or second in 5 out of 11 language-domain pairs. However, for English, Spanish, French, and Turkish, the differential with regard to the best performing system still remains large. We observe that initializing the system with general-purpose pre-trained embeddings provides a significant performance boost, which is most pronounced in the English restaurants domain.

Consequently, we hypothesize that the most straightforward way to overcome this performance differential is to initialize the system with embeddings trained on a large monolingual corpus in the target language. Incorporating domain information used by best-performing systems \cite{Pontiki2015} by pre-training on a large in-domain corpus such as the dataset released as part of the Yelp Dataset Challenge\footnote{\texttt{https://www.yelp.com/dataset\_challenge}} should further improve results.

The multi-label condition is currently enforced only after prediction through the application of a threshold, which may potentially discard promising aspects or retain erroneous ones, while the current normalization of aspect probabilities might lead to the loss of signals. To make the model more robust, the multi-label condition can be integrated more naturally into the architecture, e.g. by adapting the error function and using a trainable thresholding function as in \cite{Zhang2006}.

\subsection{Sentiment Polarity}

We report convincing results for multilingual aspect-based sentiment analysis, placing first or second for 7 out of 11 language-domain pairs. Again, the difference in performance compared to the best-performing system is largest for English, Spanish, French, and Turkish. To mitigate this, pre-trained word embeddings as described in \ref{sec:aspect_extraction} can be used.

However, without relying on dependency and constituency trees \cite{Nguyen2015a} or positional information \cite{Tang2015}, the model falls short of being able to reliably triangulate aspects, particularly in sentences with opposing sentiments toward different aspects. Simply concatenating each word vector with the aspect vector does not seem to provide the model with enough expressiveness to model truly aspect-dependent sentiment. Training the model to associate different surface forms with their aspect instantiations should help to ameliorate this.

\section{Conclusion}

In this paper, we have presented a deep learning-based approach to aspect-based sentiment analysis, which employs a convolutional neural network for aspect extraction and sentiment analysis as part of our submission to SemEval-2016 Task 5. We have demonstrated convincing results in the multilingual setting, which is particularly appropriate for neural networks due to their language and domain independence. We have evaluated our model, outlining weaknesses and potential future improvements.

\begin{table}[]
\centering
\begin{tabular}{r l | c | c | c}
\textbf{Lg.} & \textbf{Dom.} & \textbf{Acc.} & \textbf{Top Acc.} & \textbf{R.} \\\hline
EN & REST & \begin{tabular}[c]{@{}l@{}}82.072 \textbf{U}\\ 80.210 \textbf{C}\end{tabular} & 88.126 & 7/28 \\
SP & REST & 79.571 & 83.582 & 4/5 \\
FR & REST & 73.166 & 78.826 & 4/6 \\
RU & REST & 75.077 & 77.923 & 2/6 \\
DU & REST & 75.041 & 77.814 & 3/4 \\
TU & REST & 74.214 & 84.277 & 2/3 \\
AR & HOTE & 82.719 & 82.719 & 1/3 \\
EN & LAPT & \begin{tabular}[c]{@{}l@{}}78.402 \textbf{U}\\ 74.282 \textbf{C}\end{tabular} & 82.772 & 2/21 \\
DU & PHNS & 83.333 & 83.333 & 1/3 \\
CH & CAME & 78.170 & 80.457 & 2/5 \\
CH & PHNS & 72.401 & 73.346 & 2/5
\end{tabular}
\caption{Accuracy and rank of our system for aspect-based for each language and domain in comparison to the best system. For legend, refer to Table \ref{tab:aspect_extraction}.}
\label{tab:sentiment_analysis}
\end{table}

\section*{Acknowledgments}

This project has emanated from research conducted with the financial support of the Irish Research Council (IRC) under Grant Number EBPPG/2014/30 and with Aylien Ltd. as Enterprise Partner. This publication has emanated from research supported in part by a research grant from Science Foundation Ireland (SFI) under Grant Number SFI/12/RC/2289.

\bibliography{semeval_2016_task_5}
\bibliographystyle{naaclhlt2016}

\end{document}